\documentclass[11pt,a4paper]{article}
\usepackage[nohyperref]{naaclhlt2019}
\usepackage{times}
\usepackage{latexsym}
\usepackage{amsmath}
\usepackage{amssymb}
\usepackage{xcolor}
\usepackage{booktabs}
\usepackage{multirow}
\usepackage{graphicx}
\usepackage{array}
\usepackage{url}
\usepackage[normalem]{ulem}
\aclfinalcopy 
\def\aclpaperid{766} 

\title{\texttt{pair2vec}: Compositional Word-Pair Embeddings\\ for Cross-Sentence Inference}

\author{Mandar Joshi$^{\dagger}$ \qquad Eunsol Choi$^{\dagger}$ \qquad Omer Levy$^{\ddagger}$ \qquad Daniel S. Weld$^{\dagger}$ \qquad Luke Zettlemoyer$^{\dagger\ddagger}$ \\[8pt]
$^{\dagger}$ Paul G. Allen School of Computer Science \& Engineering, \\ University of Washington, Seattle, WA \\
         {\tt \{mandar90,eunsol,weld,lsz\}@cs.washington.edu}\\[8pt]
$^{\ddagger}$ Facebook AI Research, Seattle\\
        {\tt \{omerlevy,lsz\}@fb.com}\\
}

\date{}

\begin{document}

\maketitle

\begin{abstract}
Reasoning about implied relationships (e.g. paraphrastic, common sense, encyclopedic) between pairs of words is crucial for many cross-sentence inference problems. 
This paper proposes new methods for learning and using embeddings of word \emph{pairs} that implicitly represent background knowledge about such relationships. 
Our pairwise embeddings are computed as a compositional function on word representations, which is learned by maximizing the pointwise mutual information (PMI) with the contexts in which the two words co-occur.  
We add these representations to the cross-sentence attention layer of existing inference models (e.g. BiDAF for QA, ESIM for NLI), instead of extending or replacing existing word embeddings.
Experiments show a gain of 2.7\% on the recently released SQuAD 2.0 and 1.3\% on MultiNLI. Our representations also aid in better generalization with gains of around 6-7\% on adversarial SQuAD datasets, and 8.8\% on the adversarial entailment test set by ~\citet{GlocknerACL18}.
\end{abstract}

\section{Introduction}

Reasoning about 
relationships 
between pairs of words is crucial for cross sentence inference problems such as question answering (QA) and natural language inference (NLI).
In NLI, for example, given the premise ``\emph{golf is prohibitively expensive}'', inferring that the hypothesis ``\emph{golf is a cheap pastime}'' is a contradiction requires one to know that \emph{expensive} and \emph{cheap} are antonyms. 
Recent work \cite{GlocknerACL18} has shown that current models, which rely heavily on unsupervised single-word embeddings, struggle to learn such relationships.
In this paper, we show that they  
can be learned with word \emph{pair} vectors (\texttt{pair2vec}\footnote{ \url{https://github.com/mandarjoshi90/pair2vec}}), which are trained unsupervised, 
and which significantly improve performance when added to existing cross-sentence attention mechanisms.

\begin{table}
\footnotesize
\setlength{\tabcolsep}{5.5pt}
\centering

\begin{tabular}{ccc}
\toprule
\textbf{X} & \textbf{Y} & \textbf{Contexts} \\
\toprule
& & with \textbf{X} and \textbf{Y} baths \\
\textit{hot} & \textit{cold} & too \textbf{X} or too \textbf{Y} \\
& & neither \textbf{X} nor \textbf{Y} \\
\midrule
& & in \textbf{X}, \textbf{Y} \\
\textit{Portland} & \textit{Oregon} & the \textbf{X} metropolitan area in \textbf{Y} \\
& & \textbf{X} International Airport in \textbf{Y} \\
\midrule
& & food \textbf{X} are maize, \textbf{Y}, etc \\
\textit{crop} & \textit{wheat} & dry \textbf{X}, such as \textbf{Y}, \\
& & more \textbf{X} circles appeared in \textbf{Y} fields \\
\midrule
& & \textbf{X} OS comes with \textbf{Y} play \\
\textit{Android} & \textit{Google} & the \textbf{X} team at \textbf{Y} \\
& & \textbf{X} is developed by \textbf{Y} \\
\bottomrule
\end{tabular}
\caption{
Example word pairs and their contexts.
}
\label{tab:patterns}
\end{table}

Unlike single-word representations, 
which typically model 
the co-occurrence of a target word $x$ with its context $c$, our word-pair representations are learned by modeling the three-way co-occurrence between words $(x, y)$ and the context $c$ that ties them together, as seen
in Table~\ref{tab:patterns}.
While similar training signals have been used to learn models for ontology construction~\cite{Hearst92,Snow,TurneyLRA2005,ShwartzGD16} and knowledge base completion~\cite{riedel13relation}, this paper shows, for the first time, that large scale learning of pairwise embeddings can be used to directly improve the performance of neural cross-sentence inference models.

More specifically, we train a feedforward network $R(x, y)$ that learns representations for the individual words $x$ and $y$, as well as how to compose them into a single vector. 
Training is done by maximizing a generalized notion of the pointwise mutual information (PMI) among $x$, $y$, and their context $c$ using a variant of negative sampling~\cite{MikolovNips13}. 
Making $R(x,y)$ a compositional function on individual words 
alleviates 
the sparsity that necessarily comes with embedding pairs of words, even at a very large scale.

We show that our embeddings can be added to existing cross-sentence inference models, such as BiDAF++ 
\cite{SeoKFH16,ClarkACL18} for QA and ESIM~\cite{ChenQianACL17} for NLI.
Instead of changing the word embeddings that are fed into the encoder, we add the pretrained \emph{pair} representations to \emph{higher layers} in the network where cross sentence attention mechanisms are used.  This allows the model to use the background knowledge that the pair embeddings implicitly encode to 
reason about the likely relationships between the pairs of words it aligns. 

Experiments show that simply adding our word-pair embeddings to existing high-performing models, which already use ELMo \cite{PetersNAACL2018}, results in sizable gains.
We show 2.72 F1 points over the BiDAF++ model \cite{ClarkACL18} on SQuAD 2.0 
\cite{RajpurkarACL18}, as well as a 1.3 point gain over ESIM \cite{ChenQianACL17} on MultiNLI~\cite{WilliamsNAACL2018}.
Additionally, our approach generalizes well to adversarial examples, with a 6-7\% F1 increase on adversarial SQuAD \cite{jia2017adversarial} and a 8.8\% gain on the \citet{GlocknerACL18} 
NLI benchmark. 
An analysis of \texttt{pair2vec} on word analogies suggests that it complements the information in 
single-word representations, especially for 
encyclopedic and lexicographic relations.
\section{Unsupervised Pretraining}
\label{sec:pretrain}

Extending the distributional hypothesis to word pairs, we posit that similar word \emph{pairs} tend to occur in similar contexts, and that the contexts provide strong clues about the likely relationships that hold between the words (see Table~\ref{tab:patterns}). We assume a dataset of $(x, y, c)$ triplets, where each instance depicts a word pair $(x, y)$ and the context $c$ in which they appeared. 
We learn two compositional representation functions, $R(x, y)$ and $C(c)$, to encode the pair and the context, respectively, as $d$-dimensional vectors (Section~\ref{sec:representation}). The functions are trained using a variant of 
negative sampling, 
which tries to embed word pairs $(x, y)$ close to the contexts $c$ with which they appeared 
(Section~\ref{sec:objective}).

\subsection{Representation}
\label{sec:representation}

Our word-pair and context representations are both fixed-length vectors, composed from individual words. The word-pair representation function $R(x, y)$ first embeds and normalizes the individual words with a shared lookup matrix $E_a$:
\begin{equation*}
\mathbf{x} = \frac{E_a(x)}{\Vert E_a(x) \Vert } \qquad \mathbf{y} = \frac{E_a(y)}{\Vert E_a(y) \Vert }
\end{equation*}
These vectors, along with their element-wise product, are fed into a four-layer perceptron:
\begin{equation*}
R(x, y) = MLP^4(\mathbf{x}, \mathbf{y}, \mathbf{x} \circ \mathbf{y})   
\end{equation*}
The context $c = c_1 ... c_n$ is encoded as a $d$-dimensional vector using the function $C(c)$. $C(c)$ embeds each token $c_i$ with a lookup matrix $E_c$, contextualizes it with a single-layer Bi-LSTM, 
and then aggregates the entire context with attentive pooling:
\begin{align*}
    \mathbf{c}_i &= E_c(c_i) \\
    \mathbf{h_1...h_n} &=  \text{BiLSTM}(\mathbf{c}_1 ... \mathbf{c}_n) \\
    w &= \text{softmax}_i(\mathbf{k} \mathbf{h_i}) \\
    C(c) &= \sum_{i} w_i \mathbf{W} \mathbf{h}_i
\end{align*}
where $\mathbf{W} \in \mathbb{R}^{d \times d}$  and $\mathbf{k} \in \mathbb{R}^{d}$. 
All parameters, including the lookup tables $E_a$ and $E_c$, are trained.

Our representation is similar to two recently-proposed frameworks by \citet{WashioNAACL18,WashioEMNLP18}, but differs in that: (1) they use dependency paths as context, while we use surface form; (2) they encode the context as either a lookup table or the last state of a unidirectional LSTM. We also use a different objective, which we discuss next.

\subsection{Objective}
\label{sec:objective}

To optimize our 
representation functions, we consider two variants of negative sampling 
\cite{MikolovNips13}: bivariate and multivariate. The original bivariate objective models the two-way distribution of context and (monolithic) word pair co-occurrences, while our multivariate extension models the three-way distribution of word-word-context co-occurrences. 
We further 
augment the multivariate objective with typed sampling to 
upsample harder negative examples.
We discuss the impact of the bivariate and multivariate objectives (and other components) in Section \ref{sec:ablations}.

\begin{table*}[ht]
\small
\centering
\begin{tabular}{ll}
\toprule
\textbf{Bivariate}
& $J_{2NS} \left( x, y, c \right) = \log \sigma \left( R(x, y) \cdot C(c) \right) + \sum_{i=1}^{k_c} \log \sigma \left( - R(x, y) \cdot C(c^N_i) \right)$ \\ [0.15cm]
\textbf{Multivariate}
& $J_{3NS} \left( x, y, c \right) = J_{2NS} \left( x, y, c \right) + \sum_{i=1}^{k_x} \log \sigma \left( - R(x^N_i, y) \cdot C(c) \right) + \sum_{i=1}^{k_y} \log \sigma \left( - R(x, y^N_i) \cdot C(c) \right)$ \\
\bottomrule
\end{tabular}
\caption{The bivariate and multivariate negative sampling objectives. The superscript $^N$ marks randomly sampled components, with $k_{*}$ being the negative sample size per instance. The equations present per-instance objectives.}
\label{tab:objectives}
\end{table*}

\paragraph{Bivariate Negative Sampling}
Our objective aspires to make $R(x, y)$ and $C(c)$ similar (have high inner products) 
for $(x, y, c)$ that were observed together in the data.
At the same time, we wish to keep our pair vectors \emph{dis}-similar from random context vectors.
In a straightforward application of the original (bivariate) negative sampling objective, we could generate a negative example from each observed $(x, y, c)$ instance by replacing the original context $c$ with a randomly-sampled context $c^N$ (Table~\ref{tab:objectives}, $J_{2NS}$).

Assuming that the negative contexts are sampled from the empirical distribution $P(\cdot,\cdot,c)$ (with $P(x,y,c)$ being the portion of $(x,y,c)$ instances in the dataset), we can follow \citet{LevyNIPS2014} to show that this objective converges into the pointwise mutual information (PMI) between the word pair and the context. 
\begin{align*}
    R(x,y) \cdot C(c) &= \log \frac{P(x,y,c)}{k_c P(x,y,\cdot) P(\cdot,\cdot,c)}
\end{align*}
This objective mainly captures co-occurrences of monolithic pairs and contexts, and is limited by the fact that the training data, by construction, only contains pairs occurring within a sentence.
For better generalization to cross-sentence tasks, where the pair distribution differs from that of the training data, we need a multivariate objective that captures the full three-way ($x, y, c$) interaction. 

\paragraph{Multivariate Negative Sampling}
We introduce negative sampling of target words, $x$ and $y$, in addition to negative sampling of contexts $c$
(Table~\ref{tab:objectives}, $J_{3NS}$). 
Our new objective also converges to a novel multivariate generalization of PMI, different from previous PMI extensions that were inspired by information theory \cite{deCruysPMIACL2011} and heuristics \cite{JameelACL18}.\footnote{See supplementary material for their exact formulations.}
Following Levy and Goldberg (\citeyear{LevyNIPS2014}), we can show that when replacing target words in addition to contexts, our objective will converge\footnote{A full proof is provided in the supplementary material.} to the optimal value in Equation~\ref{eq:multi_pmi}: 
\begin{align}
    R(x, y) \cdot C(c) &= \log\frac{P(x, y, c)}{Z_{x,y,c} }
    \label{eq:multi_pmi}
\end{align}
where $Z_{x,y,c}$, the denominator, is:
\begin{align}
    Z_{x,y,c} 
    &= k_c P(\cdot, \cdot, c) P(x, y, \cdot) \nonumber \\
     & + k_x P(x, \cdot, \cdot) P(\cdot, y, c) \nonumber \\ 
     & + k_y P(\cdot, y, \cdot) P(x, \cdot, c) 
    \label{eq:multi_pmi_denominator}
\end{align}

This optimal value deviates from previous generalizations of PMI by having a linear mixture of marginal probability products in its denominator.
By introducing terms such as $P(x, \cdot, c)$ and $P(\cdot, y, c)$, the objective penalizes spurious correlations between words and contexts that disregard the other target word. For example, it would assign the pattern ``\emph{X is a Y}'' a high score with (\emph{banana}, \emph{fruit}), but a lower score with (\emph{cat}, \emph{fruit}).


\paragraph{Typed Sampling}
In multivariate negative sampling, we typically replace $x$ and $y$ by sampling from their unigram distributions. In addition to this, we also sample uniformly from the top 100 words according to cosine similarity using distributional word vectors. This is done to encourage the model to learn relations between specific instances as opposed to more general types. For example, using \emph{California} as a negative sample for \emph{Oregon} helps the model to learn that the pattern ``\emph{X is located in Y}'' fits the pair (\emph{Portland}, \emph{Oregon}), but not the pair (\emph{Portland}, \emph{California}). Similar adversarial constraints were used in knowledge base completion \cite{ToutanovaCPPCG15} and word embeddings \cite{LiEMNLP2017}.\footnote{Applying typed sampling also changes the value to which our objective will converge, and will replace the unigram probabilities in Equation (\ref{eq:multi_pmi_denominator}) to reflect the type-based distribution.}

\section{Integrating \texttt{pair2vec} into Models}
\label{sec:injection}

\begin{figure*}
  \centering
    \includegraphics[width=0.9\textwidth]{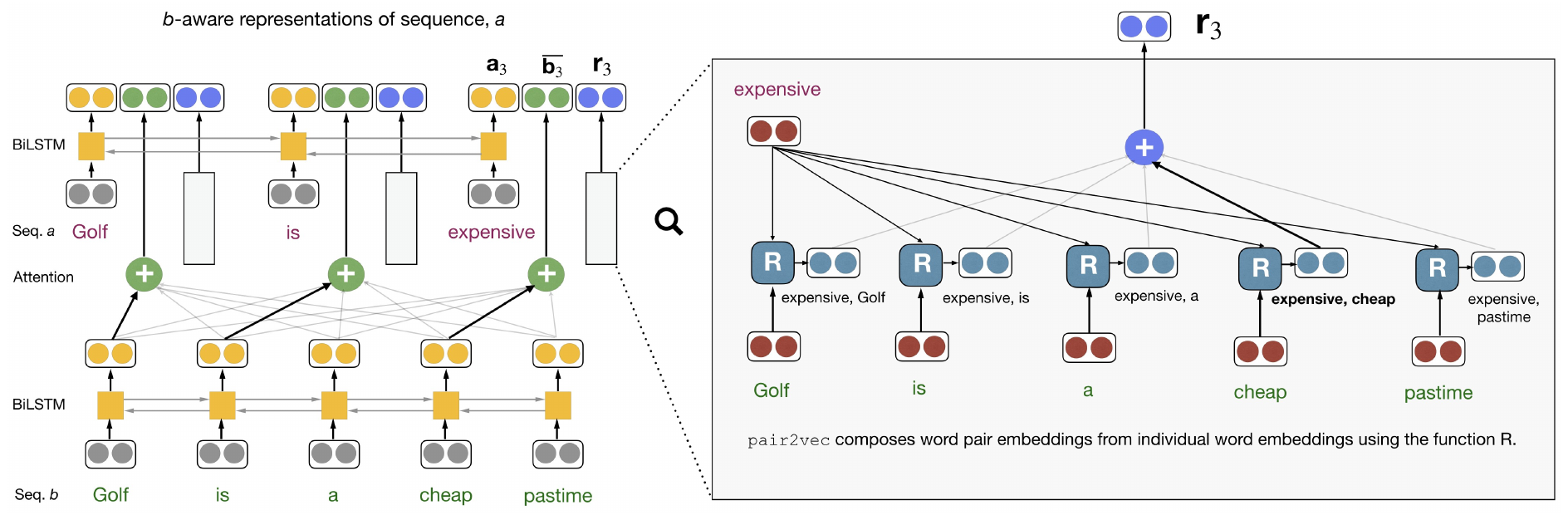}
  \caption{A typical architecture of a cross-sentence inference model (left), and how \texttt{pair2vec} is added to it (right). Given two sequences, $a$ and $b$, existing models create $b$-aware representations of words in $a$. For any word $a_i$, this typically involves the BiLSTM representation of word $a_i$ ($\mathbf{a}_i$), and an attention-weighted sum over $b$'s BiLSTM states with $a_i$ as the query ($\mathbf{\overline{b}}_i$). To these, we add the word-pair representation of $a_i$ and each word in $b$, weighted by attention ($\mathbf{r}_i$). Thicker attention arrows indicate stronger word pair alignments (e.g. \emph{cheap}, \emph{expensive}).}
  \label{fig:pair2vec_integration}
\end{figure*}

We first present a general outline for incorporating \texttt{pair2vec} into attention-based architectures, and then discuss changes made to BiDAF++ and ESIM. The key idea is to inject our pairwise representations into the attention layer by reusing the cross-sentence attention weights. In addition to attentive pooling over single word representations, we also pool over cross-sentence word pair embeddings (Figure \ref{fig:pair2vec_integration}).

\subsection{General Approach}

\paragraph{Pair Representation}
We assume that we are given two sequences $a = a_1 ... a_n$ and $b = b_1 ... b_m$. We represent the word-pair embeddings between $a$ and $b$ using the pretrained \texttt{pair2vec} model as:
\begin{equation}
\mathbf{r}_{i,j} = \left[\frac{R(a_i, b_j)}{\Vert R(a_i, b_j) \Vert}; \frac{R(b_j, a_i)}{\Vert R(b_j, a_i) \Vert}\right]
\label{eq:bi-relemb}
\end{equation}
We include embeddings in both directions, $R(a_i, b_j)$ and $R(b_j, a_i)$, because the many relations can be expressed in both directions; e.g., hypernymy can be expressed via ``\emph{X is a type of Y}'' as well as ``\emph{Y such as X}''. We take the $L_2$ normalization of each direction's pair embedding because the heavy-tailed distribution of word pairs results in significant variance of their norms.

\paragraph{Base Model}
Let $\mathbf{a}_1 ... \mathbf{a}_n$ and $\mathbf{b}_1 ... \mathbf{b}_m$ be the vector representations of sequences $a$ and $b$, as produced by the input encoder (e.g. ELMo embeddings contextualized with model-specific BiLSTMs).
Furthermore, we assume that the base model computes soft word alignments between $a$ and $b$ via co-attention (\ref{eq:wa}, \ref{eq:word_attn}), which are then used to compute $b$-aware representations of $a$:
\begin{align}
s_{i,j} &= f_{att}(\mathbf{a}_i, \mathbf{b}_j) \label{eq:wa}\\
\alpha &= \text{softmax}_j(s_{i,j}) \label{eq:word_attn} \\
\mathbf{\overline{b}}_i &= \sum_{j=0}^m \alpha_{i,j} \mathbf{b}_j \\
\mathbf{a}_i^{inf} &= \left[ \mathbf{a}_i ; \mathbf{\overline{b}}_i \right] \label{eq:inf_layer}
\end{align}
The symmetric term $\mathbf{b}_j^{inf}$ is defined analogously.
We refer to $\mathbf{a}^{inf}$ and $\mathbf{b}^{inf}$ as the inputs to the \emph{inference} layer, since this layer computes some function over aligned word pairs, typically via a feedforward network and LSTMs.
The inference layer is followed by aggregation and output layers. 
 
\paragraph{Injecting \texttt{pair2vec}}
We conjecture that the inference layer effectively learns word-pair relationships from training data, and it should, therefore, help to augment its input with \texttt{pair2vec}. We augment $\mathbf{a}_i^{inf}$ (\ref{eq:inf_layer}) with the pair vectors $\mathbf{r}_{i,j}$ (\ref{eq:bi-relemb}) by concatenating a weighted average of the pair vectors $r_{i,j}$ involving $a_i$, where the weights are the same $\alpha_{i,j}$ computed via attention in (\ref{eq:word_attn}):
\begin{align}
\mathbf{r}_i &= \sum_j \alpha_{i,j} \mathbf{r}_{i,j} \label{eq:wordpair_attn} \\
\mathbf{a}_i^{inf} &= \left[ \mathbf{a}_i ; \mathbf{\overline{b}}_i ; \mathbf{r}_i \right] 
\end{align}
The symmetric term $\mathbf{b}_j^{inf}$ is defined analogously.

\subsection{Question Answering}

We augment the inference layer in the BiDAF++ model with \texttt{pair2vec}. BiDAF++ is an improved version of the BiDAFNoAnswer \cite{SeoKFH16,LevyCoNLL2017} which includes self-attention and ELMo embeddings from ~\citet{PetersNAACL2018}. We found this variant to be stronger than the baselines presented in ~\citet{RajpurkarACL18} by over 2.5 F1. We use BiDAF++ as a baseline since its architecture is typical for QA systems, and, until recently, was state-of-the-art on SQuAD 2.0 and other benchmarks.

\paragraph{BiDAF++}
Let $\mathbf{a}$ and $\mathbf{b}$ be the outputs of the passage and question encoders respectively
(in place of the standard $\mathbf{p}$ and $\mathbf{q}$ notations). 
The inference layer's inputs $\mathbf{a}_i^{inf}$ are defined similarly to the generic model's in (\ref{eq:inf_layer}), but also contain an aggregation of the elements in $\mathbf{a}$, with better-aligned elements receiving larger weights:
\begin{align}
\mu &= \text{softmax}_i (\max_j s_{i,j}) \\
\mathbf{\hat{a}}_i &= \sum_i \mu_{i} \mathbf{a}_i \\
\mathbf{a}_i^{inf} &= \left[ \mathbf{a}_i ; \mathbf{\overline{b}}_i ; \mathbf{a}_i \circ \mathbf{\overline{b}}_i ; \mathbf{\hat{a}} \right] \label{eq:bidaf_inf}
\end{align}
In the later layers, $\mathbf{a}^{inf}$ is 
recontextualized using a BiGRU and self attention. Finally a prediction layer predicts the start and end tokens.

\paragraph{BiDAF++ with \texttt{pair2vec}}
To add our pair vectors, we simply concatenate $\mathbf{r}_i$ (\ref{eq:bi-relemb}) to $\mathbf{a}_i^{inf}$ (\ref{eq:bidaf_inf}):
\begin{align}
\mathbf{a}_i^{inf} &= \left[ \mathbf{a}_i ; \mathbf{\overline{b}}_i ; \mathbf{a}_i \circ \mathbf{\overline{b}}_i ; \mathbf{\hat{a}} ; \mathbf{r}_i \right]
\end{align}

\subsection{Natural Language Inference}

For NLI, we augment the ESIM model \cite{ChenQianACL17}, which was previously state-of-the-art on both SNLI \cite{BowmanEMNLP15} and MultiNLI \cite{WilliamsNAACL2018} benchmarks.

\paragraph{ESIM}
Let $\mathbf{a}$ and $\mathbf{b}$ be the outputs of the premise and hypothesis encoders respectively 
(in place of the standard $\mathbf{p}$ and $\mathbf{h}$ notations). 
The inference layer's inputs $\mathbf{a}_i^{inf}$ (and $\mathbf{b}_j^{inf}$) are defined similarly to the generic model's in (\ref{eq:inf_layer}):
\begin{align}
\mathbf{a}_i^{inf} &= \left[ \mathbf{a}_i ; \mathbf{\overline{b}}_i ; \mathbf{a}_i \circ \mathbf{\overline{b}}_i ; \mathbf{a}_i - \mathbf{\overline{b}}_i \right]
\label{eq:esim_inf}
\end{align}
In the later layers, $\mathbf{a}^{inf}$ and $\mathbf{b}^{inf}$ are projected, recontextualized, and converted to a fixed-length vector for each sentence using multiple pooling schemes. These vectors are then passed on to an output layer, which predicts the class.

\paragraph{ESIM with \texttt{pair2vec}}
To add our pair vectors, we simply concatenate $\mathbf{r}_i$ (\ref{eq:bi-relemb}) to $\mathbf{a}_i^{inf}$ (\ref{eq:esim_inf}):
\begin{align}
\mathbf{a}_i^{inf} &= \left[ \mathbf{a}_i ; \mathbf{\overline{b}}_i ; \mathbf{a}_i \circ \mathbf{\overline{b}}_i ; \mathbf{a}_i - \mathbf{\overline{b}}_i ; \mathbf{r}_i \right]
\label{eq:pool}
\end{align}
A similar augmentation of ESIM was recently proposed in KIM \cite{ChenACL2018}. However, their pair vectors  are composed of 
WordNet features, while our pair embeddings are learned directly from text 
(see further discussion in Section \ref{sec:related_work}).

\section{Experiments}
\label{sec:experiments}

For experiments on QA (Section \ref{sec:qa_expts}) and NLI (Section \ref{sec:nli_expts}), we use our full model which includes multivariate and typed negative sampling. We discuss ablations in Section \ref{sec:ablations}

\paragraph{Data}
We use the January 2018  dump of English Wikipedia, containing
96M sentences 
to train \texttt{pair2vec}. We restrict the vocabulary to the 100K most frequent words. Preprocessing removes all out-of-vocabulary words in the corpus. 
We consider each word pair within a window of 5 in the preprocessed corpus, and subsample\footnote{Like in \texttt{word2vec}, subsampling reduces the size of the dataset and speeds up training. For this, we define the word pair probability as the product of unigram probabilities.}
instances based on pair probability with a threshold of $5 \cdot 10^{-7}$. 
We define the context as one word each to the left and right, and all the words in between each pair, 
replacing both target words with placeholders $X$ and $Y$ (see Table~\ref{tab:patterns}). More details can be found in the supplementary material.

\subsection{Question Answering}
\label{sec:qa_expts}

\begin{table}[t]
\centering
\footnotesize
\setlength{\tabcolsep}{4pt}
\begin{tabular}{llccc}
\toprule
Benchmark & & BiDAF & + \texttt{pair2vec} & $\Delta$ \\
\midrule
\multirow{2}{*}{SQuAD 2.0} & EM & 65.66 & 68.02 & +2.36 \\
 & F1 & 68.86 & 71.58 & +2.72 \\
\midrule
\multirow{2}{*}{AddSent} & EM & 37.50 & 44.20 & +6.70 \\
 & F1 & 42.55 & 49.69 & +7.14 \\
\midrule
\multirow{2}{*}{AddOneSent} & EM & 48.20 & 53.30 & +5.10 \\
& F1 & 54.02 & 60.13 & +6.11 \\
\bottomrule
\end{tabular}
\caption{Performance on SQuAD 2.0 and adversarial SQuAD (AddSent and AddOneSent) benchmarks, with and without \texttt{pair2vec}. All models have ELMo.}
\label{tab:qa_res}
\end{table}

We experiment on the SQuAD 2.0 QA benchmark \cite{RajpurkarACL18}, as well as the adversarial datasets of SQuAD 1.1 \cite{Rajpurkar:16,jia2017adversarial}. Table~\ref{tab:qa_res} shows the performance of BiDAF++, with ELMo , before and after adding \texttt{pair2vec}. 
Experiments on SQuAD 2.0 show that our pair representations improve performance by 2.72 F1.
Moreover, adding \texttt{pair2vec} also results in better generalization on the adversarial SQuAD datasets with gains of 7.14 and 6.11 F1.

\subsection{Natural Language Inference}
\label{sec:nli_expts}

\begin{table}[t]
\footnotesize
\centering
\setlength{\tabcolsep}{4pt}
\begin{tabular}{lccc}
\toprule
Benchmark & ESIM & + \texttt{pair2vec} & $\Delta$ \\
\midrule
Matched & 79.68 & 81.03 & +1.35 \\
Mismatched & 78.80 & 80.12 & +1.32 \\
\bottomrule
\end{tabular}
\caption{Performance on MultiNLI, with and without \texttt{pair2vec}. All models have ELMo.}
\label{tab:nli_res}
\end{table}

\begin{table}[t]
\footnotesize
\centering
\setlength{\tabcolsep}{4pt}
\begin{tabular}{lc}
\toprule
Model & Accuracy \\
\midrule
\textbf{Rule-based Models} & \\
WordNet Baseline & 85.8 \\
\midrule
\textbf{Models with GloVe} & \\
ESIM \cite{ChenQianACL17} & 77.0 \\ 
KIM \cite{ChenACL2018} & 87.7 \\
ESIM + \texttt{pair2vec} & \textbf{92.9} \\
\midrule
\textbf{Models with ELMo} & \\
ESIM \cite{PetersNAACL2018} & 84.6 \\
ESIM + \texttt{pair2vec} & \textbf{93.4} \\
\bottomrule
\end{tabular}
\caption{Performance on the adversarial NLI test set of \citet{GlocknerACL18}. }
\label{tab:glockner_res}
\end{table}



We report the performance of our model on MultiNLI and the adversarial test set from \citet{GlocknerACL18} in Table \ref{tab:glockner_res}. We outperform the ESIM + ELMo baseline by 1.3\% on the matched and mismatched portions of the dataset. 

We also record a gain of 8.8\% absolute over ESIM on the \citet{GlocknerACL18} dataset, setting a new state of the art. Following standard practice~\cite{GlocknerACL18}, we train all models on a combination of SNLI \cite{BowmanEMNLP15} and MultiNLI. \citet{GlocknerACL18} show that with the exception of KIM \cite{ChenACL2018}, which uses WordNet features, several NLI models fail to generalize to this setting which involves lexical inference. For a fair comparison with KIM on the Glockner test set, we replace ELMo with GLoVE embeddings, and still outperform KIM by almost halving the error rate.

\subsection{Ablations}
\label{sec:ablations}

\begin{table}[t]
\centering
\footnotesize
\begin{tabular}{lcc}
\toprule
Model & EM ($\Delta$) & F1 ($\Delta$) \\
\midrule
\texttt{pair2vec} (Full Model) & 69.20~~~~~~~~~~~~ & 72.68~~~~~~~~~~~~ \\
\midrule
Composition: 2 Layers & 68.35 (-0.85) & 71.65 (-1.03) \\
Composition: Multiply & 67.10 (-2.20) & 70.20 (-2.48) \\
Objective: Bivariate NS & 68.63 (-0.57) & 71.98 (-0.70) \\
Unsupervised: Pair Dist & 67.07 (-2.13) & 70.24 (-2.44) \\
\midrule
No \texttt{pair2vec} (BiDAF) & 66.66 (-2.54) & 69.90 (-2.78) \\
\bottomrule
\end{tabular}
\caption{Ablations on the Squad 2.0 development set show that argument sampling as well as using a deeper composition function are useful.}
\label{tab:ablations}
\end{table}

Ablating parts of \texttt{pair2vec} shows that all components of the model (Section~\ref{sec:pretrain}) are useful. We ablate each component and report the EM and F1 on the development set of SQuAD 2.0 (Table \ref{tab:ablations}). The full model, which uses a 4-layer MLP for $R(x,y)$ and trains with multivariate negative sampling, achieves the highest F1 of 72.68. 

We experiment with two alternative composition functions, a 2-layer MLP (\textit{Composition: 2 Layers}) and element-wise multiplication (\textit{Composition: Multiply}), which yield significantly smaller gains over the baseline BiDAF++ model. This demonstrates the need for a deep composition function. Eliminating sampling of target words $(x, y)$ from the objective (\textit{Objective: Bivariate NS}) results in a drop of 0.7 F1, accounting for about a quarter of the overall gain. This suggests that while the bulk of the signal is mined from the pair-context interactions, there is also valuable information in other interactions as well.

We also test whether specific pre-training of word \emph{pair} representations is useful by replacing \texttt{pair2vec} embeddings with the vector offsets of pre-trained word embeddings (\textit{Unsupervised: Pair Dist}). We follow the PairDistance method for word analogies \cite{MikolovNAACL13}, and represent the pair $(x,y)$ as the L2 normalized difference of single-word vectors:
$(\mathbf{x}-\mathbf{y}) / \Vert \mathbf{x}-\mathbf{y} \Vert$.
We use the same fastText ~\cite{bojanowski2017enriching} word vectors with which we initialized \texttt{pair2vec} before training. We observe a gain of only 0.34 F1 over the baseline.


\section{Analysis} 

\begin{figure}
  \centering
    \includegraphics[width=0.45\textwidth]{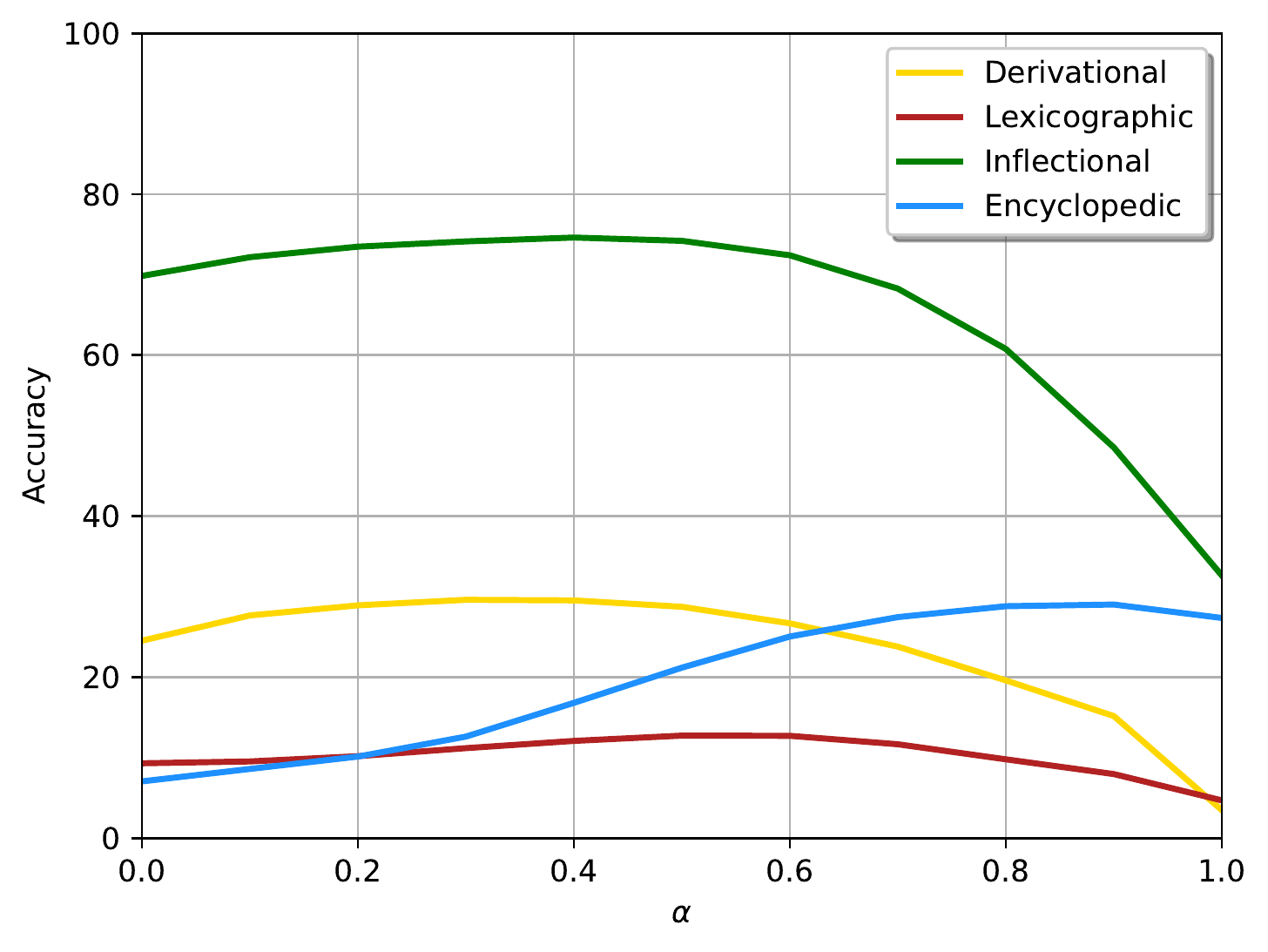}
  \caption{Accuracy as a function of the interpolation parameter $\alpha$ (see Eq. (\ref{eq:analogy_score})). The $\alpha$=0 configuration relies only on fastText ~\cite{bojanowski2017enriching}, while $\alpha$=1 reflects \texttt{pair2vec}.}
  \label{fig:dlie}
\end{figure}

\begin{table}[t]
\footnotesize
\centering
\begin{tabular}{lccc}
\toprule
Relation & \texttt{3CosAdd} & +\texttt{pair2vec} & $\alpha^*$ \\
\midrule
Country:Capital & ~~1.2 & 86.1 & 0.9 \\
Name:Occupation & ~~1.8 & 44.6 & 0.8 \\
Name:Nationality & ~~0.1 & 42.0 & 0.9 \\
UK City:County & ~~0.7 & 31.7 & 1.0 \\
Country:Language & ~~4.0 & 28.4 & 0.8 \\
Verb 3pSg:Ved & 49.1 & 61.7 & 0.6 \\
Verb Ving:Ved & 61.1 & 73.3 & 0.5 \\
Verb Inf:Ved & 58.5 & 70.1 & 0.5 \\
Noun+less & ~~4.8 & 16.0 & 0.2 \\
Substance Meronym & ~~3.8 & 14.5 & 0.6 \\
\bottomrule
\end{tabular}
\caption{The top 10 analogy relations for which interpolating with \texttt{pair2vec} improves performance.  $\alpha^*$ is the optimal interpolation parameter for each relation.} 
\label{tab:top_relations}
\end{table}

\begin{table*}[t]
\footnotesize
\centering
\begin{tabular}
{cccl}
\toprule
\textbf{Relation} & \textbf{Context} & \textbf{X} & \textbf{Y (Top 3)} \\
\midrule
\multirow{2}{*}{Antonymy/Exclusion} & \multirow{2}{*}{either X or Y} & accept & \emph{reject}, \emph{refuse}, recognise \\
 & & hard & \emph{soft}, \emph{brittle}, \emph{polished} \\
\midrule 
\multirow{2}{*}{Hypernymy} & \multirow{2}{*}{including X and other Y} & copper & ones, \emph{metals}, \emph{mines} \\
& & google & apps, \emph{browsers}, \emph{searches} \\
\midrule
\multirow{2}{*}{Hyponymy} & \multirow{2}{*}{X like Y} & cities & solaris, \emph{speyer}, \emph{medina} \\
& & browsers & \emph{chrome}, \emph{firefox}, \emph{netscape} \\
\midrule
\multirow{2}{*}{Co-hyponymy} & \multirow{2}{*}{, X , Y ,} & copper & \emph{malachite}, \emph{flint}, ivory \\
& & google & \emph{microsoft}, \emph{bing}, \emph{yahoo} \\
\midrule
\multirow{2}{*}{City-State} & \multirow{2}{*}{in X , Y .} & portland & \emph{oregon}, \emph{maine}, \textit{dorset} \\
& & dallas & \emph{tx}, \emph{texas}, va\\
\midrule
\multirow{2}{*}{City-City} & \multirow{2}{*}{from X to Y .} & portland & \emph{salem}, \emph{astoria}, ogdensburg \\
& & dallas & \emph{denton}, \emph{allatoona}, \emph{addison} \\
\midrule
\multirow{2}{*}{Profession} & \multirow{2}{*}{X , a famous Y ,} & ronaldo & \emph{footballer}, portuguese, \emph{player} \\
& & monet & \emph{painter}, painting, butterfly \\
\bottomrule
\end{tabular}
\caption{Given a context $c$ and a word $x$, we select the top 3 words $y$ from the entire vocabulary using our scoring function $R(x,y) \cdot C(c)$. The analysis suggests that the model tends to rank correct matches (italics) over others.} 
\label{tab:kbc_word_prediction}
\end{table*}

In Section~\ref{sec:experiments}, we showed that \texttt{pair2vec} adds information complementary to single-word representations like ELMo. Here, we ask what this extra information is, and try to characterize which word relations are better captured by \texttt{pair2vec}. To that end, we evaluate performance on a word analogy dataset with over 40 different relation types (Section \ref{sec:quantitative_analysis}), and observe how \texttt{pair2vec} fills hand-crafted relation patterns (Section \ref{sec:qualitative_analysis}).

\subsection{Quantitative Analysis: Word Analogies}
\label{sec:quantitative_analysis}

\paragraph{Word Analogy Dataset}
Given a word pair $(a, b)$ and word $x$, the word analogy task involves predicting a word $y$ such that $a:b::x:y$.
We use the Bigger Analogy Test Set
\citep[BATS,][]{GladkovaBATS2016} which contains four groups of relations: encyclopedic semantics (e.g., person-profession as in \textit{Einstein}-\textit{physicist}), lexicographic semantics (e.g., antonymy as in \textit{cheap}-\textit{expensive}), derivational morphology (e.g., noun forms as in \textit{oblige}-\textit{obligation}), and inflectional morphology (e.g., noun-plural as in \textit{bird}-\textit{birds}). Each group contains 10 sub-relations.

\paragraph{Method}
We interpolate \texttt{pair2vec} and \texttt{3CosAdd} \cite{MikolovNAACL13,LevyCoNLL2014} scores on fastText  embeddings, as follows:
\begin{align}
    \textrm{score}(y) &= \alpha \cdot   \textrm{cos} (\mathbf{r}_{a,b}, \mathbf{r}_{x,y}) \nonumber \\  
    &+ (1-\alpha) \cdot \textrm{cos}(\mathbf{b} - \mathbf{a} + \mathbf{x}, \mathbf{y})
    \label{eq:analogy_score}
\end{align}
where $\mathbf{a}$, $\mathbf{b}$, $\mathbf{x}$, and $\mathbf{y}$ represent fastText embeddings\footnote{The fastText embeddings in the analysis were retrained using the same Wikipedia corpus used to train \texttt{pair2vec} to control for the corpus when comparing the two methods.} and 
$\mathbf{r}_{a, b}$, $\mathbf{r}_{x, y}$ represent the $\texttt{pair2vec}$ embedding for the word pairs 
$(a, b)$ and $(x, y)$, respectively; $\alpha$ is the linear interpolation parameter. 
Following prior work~\cite{MikolovNAACL13}, we return the highest-scoring $y$ in the entire vocabulary, excluding the given words $a$, $b$, and $x$.

\paragraph{Results}
Figure \ref{fig:dlie} shows how the accuracy on each category of relations varies with $\alpha$.
For all four groups, adding \texttt{pair2vec} to \texttt{3CosAdd} results in significant gains. In particular, the biggest relative improvements are observed for encyclopedic (356\%) and lexicographic (51\%) relations.

Table~\ref{tab:top_relations} shows the specific relations in which \texttt{pair2vec} made the largest absolute impact.
The gains are particularly significant for relations where fastText embeddings provide limited signal. For example, the accuracy for \emph{substance meronyms} goes from 3.8\% to 14.5\%. 
In some cases, there is also a synergistic effect; for instance, in \emph{noun+less}, \texttt{pair2vec} alone scored 0\% accuracy, but mixing it with \texttt{3CosAdd}, which got 4.8\% on its own, yielded 16\% accuracy.

These results, alongside our experiments in Section~\ref{sec:experiments}, strongly suggest that \texttt{pair2vec} encodes information complementary to that in single-word embedding methods such as fastText and ELMo.


\subsection{Qualitative Analysis: Slot Filling}
\label{sec:qualitative_analysis}

To further explore how \texttt{pair2vec} encodes such complementary information, we consider a setting similar to that of knowledge base completion: given a Hearst-like context pattern $c$ and a single word $x$, predict the other word $y$ from the entire vocabulary. We rank candidate words $y$ based on the scoring function in our training objective: $R(x,y) \cdot C(c)$.
We use a fixed set of example relations and manually define their predictive context patterns and a small set of candidate words $x$.

Table~\ref{tab:kbc_word_prediction} shows the top three $y$ words. The model embeds 
$(x,y)$ pairs close to contexts that reflect their relationship. 
For example, substituting \emph{Portland} in the city-state pattern (``\emph{in X, Y.}''), the top two words are \emph{Oregon} and \emph{Maine}, both US states with cities named Portland. When used with the city-city pattern (``\emph{from X to Y.}''), the top two words are \emph{Salem} and \emph{Astoria}, both cities in Oregon. The word-context interaction often captures multiple relations; for example, \emph{Monet} is used to refer to the painter (\emph{profession}) as well as his paintings. 

As intended, \texttt{pair2vec} captures the three-way word-word-context interaction, and not just the two-way word-context interaction (as in single-word embeddings). This profound difference allows \texttt{pair2vec} to complement single-word embeddings with additional information.



\section{Related Work}
\label{sec:related_work}

\paragraph{Pretrained Word Embeddings}
Many state-of-the-art models initialize their word representations using pretrained embeddings such as \texttt{word2vec} \cite{MikolovNips13} or ELMo \cite{PetersNAACL2018}. These representations are typically trained using an interpretation of the Distributional Hypothesis \cite{Harris1954} in which the bivariate distribution of target words and contexts is modeled. Our work deviates from the word embedding literature in two major aspects. First, our goal is to represent word \emph{pairs}, not individual words. Second, our new PMI formulation models the \emph{trivariate} word-word-context distribution. Experiments show that our pair embeddings can complement single-word embeddings.

\paragraph{Mining Textual Patterns}
There is extensive literature on mining textual patterns to predict relations between words \cite{Hearst92,Snow,TurneyLRA2005,riedel13relation,cruys2014selectional,ToutanovaCPPCG15,Shwartz2016PathbasedVD}.
These approaches focus mostly on relations between pairs of nouns (perhaps with the exception of VerbOcean \cite{ChklovskiEMNLP2004}). 
More recently, they have been expanded to predict relations between unrestricted pairs of words \cite{JameelACL18,EspinosaCOLING2018}, assuming that each word-pair was observed together during pretraining. 
\citet{WashioNAACL18,WashioEMNLP18} relax this assumption with a compositional model that can represent any pair, as long as each word appeared (individually) in the corpus.

These methods are evaluated on either intrinsic relation prediction tasks, such as BLESS \cite{Baroni2011BLESS} and CogALex \cite{SantusCOLING16}, or knowledge-base population benchmarks, e.g. FB15 \cite{BordesNIPS2013}. To the best of our knowledge, our work is the first to integrate pattern-based methods into modern high-performing semantic models and evaluate their impact on complex end-tasks like QA and NLI.

\paragraph{Integrating Knowledge in Complex Models}
\citet{Ahn2016ANK} integrate Freebase facts into a language model using a copying mechanism over fact attributes.  \citet{Yang2017LeveragingKB} modify the LSTM cell to incorporate WordNet and NELL knowledge for event and entity extraction. 
For cross-sentence inference tasks, \citet{Weissenborn:18}, \citet{Bauer2018CS}, and \citet{MihaylovACL18} dynamically refine word representations by reading assertions from ConceptNet and Wikipedia abstracts. Our approach, on the other hand, relies on a relatively simple extension of existing cross-sentence inference models. Furthermore, we do not need to dynamically retrieve and process knowledge base facts or Wikipedia texts, and just pretrain our pair vectors in advance.

KIM \cite{ChenQianACL17} integrates word-pair vectors into the ESIM model for NLI in a very similar way to ours. However, KIM's word-pair vectors contain only hand-engineered word-relation indicators from WordNet, whereas our word-pair vectors are automatically learned from unlabeled text. Our vectors can therefore reflect relation types that do not exist in WordNet (such as \emph{profession}) as well as word pairs that do not have a direct link in WordNet (e.g. \emph{bronze} and \emph{statue}); see Table~\ref{tab:kbc_word_prediction} for additional examples.

\section{Conclusion and Future Work}
We presented new methods for training and using word \emph{pair} embeddings that implicitly represent background knowledge. Our pair embeddings are computed as a compositional function of the individual word representations, which is learned by maximizing a variant of the PMI with the contexts in which the the two words co-occur.  Experiments on cross-sentence inference benchmarks demonstrated that adding these representations to existing models results in sizable improvements for both in-domain and adversarial settings.

Published concurrently with this paper, BERT ~\cite{devlin2018bert}, which uses a masked language model objective, has reported dramatic gains on multiple semantic benchmarks including question-answering, natural language inference, and named entity recognition. Potential avenues for future work include multitasking BERT with \texttt{pair2vec} in order to more directly incorporate reasoning about word pair relations into the BERT objective.

\section*{Acknowledgments}
We would like to thank Anna Rogers (Gladkova), Qian Chen, Koki Washio, Pranav Rajpurkar, and Robin Jia for their help with the evaluation. We are also grateful to members of the UW and FAIR NLP groups, and anonymous reviewers for their thoughtful comments and suggestions.

\bibliography{ref}
\bibliographystyle{acl_natbib}

\clearpage
\appendix
\section{Implementation Details}
\paragraph{Hyperparameters}
For both word pairs and contexts, we use 300-dimensional word embeddings initialized with FastText \cite{bojanowski2017enriching}. The context representation uses a single-layer Bi-LSTM with a hidden layer size of 100. We use 2 negative context samples and 3 negative argument samples for each pair-context tuple.

For pre-training, we used stochastic gradient descent with an initial learning rate of 0.01. We reduce the learning rate by a factor of 0.9 if the loss does not decrease for 300K steps. We use a batch size of 600, and train for 12 epochs.\footnote{On Titan X GPUs, the training takes about a week.}

For both end-task models, we use AllenNLP's implementations \cite{Gardner2017AllenNLP} with default hyperparameters; we did not change any setting before or after injecting \texttt{pair2vec}. We use 0.15 dropout on our pretrained pair embeddings.

\section{Multivariate Negative Sampling}
\label{appendix_proof}
In this appendix, we elaborate on mathematical details of multivariate negative sampling to support our claims in Section~\ref{sec:objective}.

\subsection{Global Objective}
\label{sec:global_objective}

Equation~(Table~\ref{tab:objectives}, $J_{3NS}$) in Section~\ref{sec:objective} characterizes the local objective for each data instance. To understand the mathematical properties of this objective, we must first describe the global objective in terms of the entire dataset. However, this cannot be done by simply summing the local objective for each $(x, y, c)$, since each such example may appear multiple times in our dataset. Moreover, due to the nature of negative sampling, the number of times an $(x, y, c)$ triplet appears as a positive example will almost always be different from the number of times it appears as a negative one. Therefore, we must determine the frequency in which each triplet appears in each role.

We first denote the number of times the example $(x, y, c)$ appears in the dataset as $\#(x, y, c)$; this is also the number of times $(x, y, c)$ is used as a positive example. We observe that the expected number of times $(x, y, c)$ is used as a corrupt $x$ example is $k_x P(x, \cdot, \cdot) \#(\cdot, y, c)$, since $(x, y, c)$ can only be created as a corrupt $x$ example by randomly sampling $x$ from an example that already contained $y$ and $c$. The number of times $(x, y, c)$ is used as a corrupt $y$ or $c$ example can be derived analogously. Therefore, the global objective of our trenary negative sampling approach is:
\begin{align}
    J_{3NS}^{\text{Global}} &= \sum_{(x,y,c)} J_{3NS}^{x,y,c} \label{eq:global_objective} \\
    J_{3NS}^{x,y,c} &= \#(x,y,c) \cdot \log \sigma \left( S_{x,y,c} \right) \nonumber \\
    &+ Z_{x,y,c}' \cdot \log \sigma \left( - S_{x,y,c} \right) \label{eq:element_objective} \\
    Z_{x,y,c}' &= k_x P(x, \cdot, \cdot) \#(\cdot, y, c) J_-^{x,y,c} \nonumber \\
    &+ k_y P(\cdot, y, \cdot) \#(x, \cdot, c) J_-^{x,y,c} \nonumber \\
    &+ k_c P(\cdot, \cdot, c) \#(x, y, \cdot) J_-^{x,y,c} \label{eq:denominator} \\
    S_{x,y,c} &= R(x, y) \cdot C(c) \label{eq:score}
\end{align}

\begin{table*}[t]
\small
\centering
\begin{tabular}{lcc}
\toprule
\multirow{4}{*}{\cite{deCruysPMIACL2011}} & \multirow{2}{*}{$SI_1 (x, y, c)$} & \multirow{2}{*}{$\log\frac{P(x, y, \cdot) P(x, \cdot, c) P(\cdot, y, c)}{P(x, \cdot, \cdot)P(\cdot, y, \cdot)P(\cdot, \cdot, c)P(x, y, c)}$} \\
& & \\
& \multirow{2}{*}{$SI_2 (x, y, c)$} & \multirow{2}{*}{$\log\frac{P(x, y, c)}{P(x, \cdot, \cdot)P(\cdot, y, \cdot)P(\cdot, \cdot, c)}$} \\
& & \\
\midrule
\multirow{4}{*}{\cite{JameelACL18}} & \multirow{2}{*}{$SI_3 (x, y, c)$} & \multirow{2}{*}{$\log\frac{P(x, y, c)}{P(x, y, \cdot)P(\cdot, \cdot, c)}$} \\
& & \\
& \multirow{2}{*}{$SI_4 (x, y, c)$} & \multirow{2}{*}{$\log\frac{P(x, y, c)P(\cdot, \cdot, c)}{P(x, \cdot, c)P(\cdot, y, c)}$} \\
& & \\
\midrule
\multirow{2}{*}{(This Work)} & \multirow{2}{*}{$MVPMI (x, y, c)$} & \multirow{2}{*}{$\log\frac{P(x, y, c)}{k_c P(\cdot, \cdot, c) P(x, y, \cdot) + k_x P(x, \cdot, \cdot) P(\cdot, y, c) + k_y P(\cdot, y, \cdot) P(x, \cdot, c)}$} \\
& & \\
\bottomrule
\end{tabular}
\caption{Multivariate generalizations of PMI.}
\label{tab:pmi_generalizations}
\end{table*}

\subsection{Relation to Multivariate PMI}
\label{sec:pmi_proof}

With the global objective, we can now ask what is the optimal value of $S_{x,y,c}$ (\ref{eq:score}) by comparing the partial derivative of (\ref{eq:global_objective}) to zero. This derivative is in fact equal to the partial derivative of (\ref{eq:element_objective}), since it is the only component of the global objective in which $R(x, y) \cdot C(c)$ appears:
\begin{align*}
0 = \frac{\partial J_{3NS}^{\text{Global}}}{\partial S_{x,y,c}} = \frac{\partial J_{3NS}^{x,y,c}}{\partial S_{x,y,c}}
\end{align*}
The partial derivative of (\ref{eq:element_objective}) can be expressed as:
\begin{align*}
0 = \#(x,y,c) \cdot \sigma \left(- S_{x,y,c} \right) - Z_{x,y,c}' \cdot \sigma \left(S_{x,y,c} \right)
\end{align*}
which can be reformulated as:
\begin{align*}
S_{x,y,c} = \log\frac{\#(x,y,c)}{Z_{x,y,c}'}
\end{align*}
By expanding the fraction by $1 / \#(\cdot,\cdot,\cdot)$ (i.e. dividing by the size of the dataset), we essentially convert all the frequency counts (e.g. $\#(x,y,z)$) to empirical probabilities (e.g. $P(x,y,z)$), and arrive at Equation~(\ref{eq:multi_pmi}) in Section~\ref{sec:objective}.

\subsection{Other Multivariate PMI Formulations}
\label{sec:other_pmi}

Previous work has proposed different multivariate formulations of PMI, shown in Table~\ref{tab:pmi_generalizations}. \citet{deCruysPMIACL2011} presented specific interaction information ($SI_1$) and specific correlation ($SI_2$). In addition to those metrics, \citet{JameelACL18} experimented with $SI_3$, which is the bivariate PMI between $(x, y)$ and $c$, and with $SI_4$. Our formulation deviates from previous work, and, to the best of our knowledge, cannot be trivially expressed by one of the existing metrics.

\end{document}